\documentclass[runningheads]{llncs}
\usepackage[T1]{fontenc}
\usepackage{graphicx}
\usepackage{amsmath}
\usepackage{booktabs} 
\begin{document}
\title{Effect of Data Augmentation on Conformal Prediction for Diabetic Retinopathy}

\titlerunning{Effect of Data Augmentation on CP for DR}
\author{%
  Rizwan Ahamed\inst{1}
  \and
  Annahita Amireskandari\inst{1}
  \and
  Joel Palko\inst{1}
  \and
  Carol Laxson\inst{1}
  \and
  Binod Bhattarai\inst{2}
  \and
  Prashnna Gyawali\inst{1}
}
\authorrunning{Ahamed et al.}

\institute{%
  West Virginia University, Morgantown, WV 26506, USA\\
  \and
  University of Aberdeen, Aberdeen AB24 3FX, UK\\
}
\maketitle
\begin{abstract}
The clinical deployment of deep learning models for high-stakes tasks such as diabetic retinopathy (DR) grading requires demonstrable reliability. While models achieve high accuracy, their clinical utility is limited by a lack of robust uncertainty quantification. Conformal prediction (CP) offers a distribution-free framework to generate prediction sets with statistical guarantees of coverage. However, the interaction between standard training practices like data augmentation and the validity of these guarantees is not well understood. In this study, we systematically investigate how different data augmentation strategies affect the performance of conformal predictors for DR grading. Using the DDR dataset, we evaluate two backbone architectures—ResNet-50 and a Co-Scale Conv-Attentional Transformer (CoaT)—trained under five augmentation regimes: no augmentation, standard geometric transforms, CLAHE, Mixup, and CutMix. We analyze the downstream effects on conformal metrics, including empirical coverage, average prediction set size, and correct efficiency. Our results demonstrate that sample-mixing strategies like Mixup and CutMix not only improve predictive accuracy but also yield more reliable and efficient uncertainty estimates. Conversely, methods like CLAHE can negatively impact model certainty. These findings highlight the need to co-design augmentation strategies with downstream uncertainty quantification in mind to build genuinely trustworthy AI systems for medical imaging.

\end{abstract}
\section{Introduction}

The integration of deep learning (DL) into medical image analysis has driven a paradigm shift in computer-aided diagnosis. In ophthalmology, DL models, particularly Convolutional Neural Networks (CNNs) and novel hybrid architectures, have demonstrated remarkable capabilities in detecting and classifying conditions like DR from fundus images, with precision that often rivals trained ophthalmologists \cite{gulshan2016development,huang2023identifying,liu2024lesion}. As a leading cause of preventable blindness, the early and accurate severity grading of DR is essential for timely intervention \cite{ting2016diabetic}. Automated systems thus hold immense promise for scaling screening programs and improving access to care.

However, the transition of these high-performing models from research studies to clinical workflows is hindered by a significant challenge: a lack of demonstrable reliability and trust \cite{kelly2019key}. 
This need for reliability has spurred research into Uncertainty Quantification (UQ). UQ methods aim to equip models with the ability to measure their confidence. Common approaches include Bayesian neural networks, which learn a distribution over model weights \cite{gal2016dropout}, and deep ensembles, which aggregate predictions from multiple models \cite{lakshminarayanan2017simple}. While powerful, these methods can be computationally intensive and rely on strong distributional assumptions.

In this context, CP has emerged as an advantageous UQ framework due to its mathematical rigor and distribution-free nature \cite{vovk2005algorithmic,shafer2008tutorial}. Instead of a single-point prediction, CP outputs a \textit{prediction set}—a subset of possible labels that is guaranteed to contain the true label with a predefined probability (e.g., 90\%). This property holds without assumptions about the underlying data distribution or model architecture, provided the data are exchangeable \cite{angelopoulos2021gentle}. The clinical implications are profound: a small prediction set signals high confidence, while a large set indicates high uncertainty and can automatically flag a case for expert review \cite{angelopoulos2024conformal}.

Concurrently, data augmentation is an indispensable technique for training state-of-the-art models, especially in medical imaging \cite{shorten2019survey}. By applying transformations ranging from geometric operations to sample-mixing strategies like Mixup \cite{zhang2018mixupempiricalriskminimization} and CutMix \cite{yun2019cutmix}, practitioners significantly improve model generalization. This standard practice, however, introduces a critical and largely unexamined tension with CP. Data augmentation inherently alters the training data distribution, potentially violating the exchangeability assumption that underpins CP's statistical guarantees \cite{smucny2022data}. An augmentation strategy that boosts predictive accuracy might, in turn, compromise the validity of the uncertainty estimates. 

To this end, our paper presents a systematic investigation into this trade-off, aiming to provide crucial insights for building reliable AI systems for medical diagnosis. We analyze two popular architectures and five different augmentation schemes on the Diabetic Retinopathy dataset and found notable differences in augmentation strategies when evaluated using conformal-based metrics.

\section{Background}
\subsection{Conformal Prediction}

Conformal Prediction is a powerful framework that directly addresses uncertainty by providing statistically rigorous, distribution-free guarantees on its predictions \cite{vovk2005algorithmic,shafer2008tutorial,zhou2025conformal}. A critical component of CP is the \textit{nonconformity measure}, $A$, a function that scores how "atypical" a data point $(x, y)$ is. For a set of calibration examples $\{(x_1, y_1), ..., (x_n, y_n)\}$ and a new test example $x_{n+1}$, we can calculate a nonconformity score $\alpha_i = A(x_i, y_i)$ for each calibration point. To determine the plausibility of a potential label $\hat{y}$ for the test point, we compute its score $\alpha_{n+1} = A(x_{n+1}, \hat{y})$. The p-value for $\hat{y}$ is then the fraction of calibration scores that are at least as large as the test score:
\begin{equation}
p(\hat{y}) = \frac{|\{i=1,...,n : \alpha_i \ge \alpha_{n+1}\}| + 1}{n + 1}.
\end{equation}
Given a user-specified significance level $\epsilon \in (0, 1)$, a prediction set $\Gamma^{\epsilon}$ is formed by including all labels whose p-values exceed this threshold:
\begin{equation}
\Gamma^{\epsilon} := \{ \hat{y} : p(\hat{y}) > \epsilon \}.
\end{equation}
The core strength of CP is that this prediction set is guaranteed to satisfy $P(y_{true} \in \Gamma^{\epsilon}) \ge 1-\epsilon$. This property, known as marginal coverage, is a powerful tool for risk management in clinical settings \cite{vazquez2022conformal}.

However, one of the limitation of CP framework is its reliance on the mathematical assumption of exchangeability \cite{barber2023conformal}. This assumption requires that the joint distribution of the data is invariant to permutation. If the training, calibration, and test data are not exchangeable—due to factors like distribution drift or, critically, data augmentation—the coverage guarantee of CP is no longer valid.

\subsection{Data Augmentation}

Data augmentation is a cornerstone of modern deep learning, serving as a powerful regularization technique to improve model generalization and robustness, particularly when training data is scarce \cite{shorten2019survey}. This is especially relevant in medical imaging, in tasks like DR classification, where datasets are often imbalanced. Augmentation, from basic affine transformations to the generation of synthetic images \cite{srilakshmi2025data}, is essential for training robust models.

Recent work has also demonstrated the effectiveness of sample-mixing techniques, such as Mixup \cite{zhang2018mixupempiricalriskminimization} and CutMix \cite{yun2019cutmix}. These methods create new training samples by interpolating or combining patches from existing images and their corresponding labels. While highly effective at improving raw predictive accuracy, these methods fundamentally alter the training data distribution by creating synthetic, out-of-distribution examples. This practice creates a direct conflict with the exchangeability assumption central to CP, motivating our investigation into the trade-off between model accuracy and the reliability of uncertainty estimates.

\section{Methodology}

\subsection{Dataset and Preprocessing}

We conducted all experiments using the publicly available DDR dataset \cite{li2019diagnostic}. This dataset contains high-resolution retinal fundus photographs annotated with DR grades from 0 (no DR) to 4 (proliferative DR), following the international clinical DR severity scale. The images were preprocessed by resizing to $224 \times 224$ pixels to match the input requirements of the models.
For the CP procedure, the dataset was divided into three distinct, non-overlapping sets: a \textbf{proper training set} (10,017 images), a \textbf{calibration set} (2,004 images), and a \textbf{test set} (501 images). This split was performed using patient-level, stratified sampling to preserve the class distribution across DR grades and prevent data leakage from the same patient appearing in multiple subsets.

\subsection{Model Architectures}

We compared two widely-used deep learning architectures for our analysis: a standard CNN and a modern hybrid-attention model.

\paragraph{ResNet-50:} As a representative CNN, we used the ResNet-50 architecture \cite{he2016deep}, a 50-layer deep residual network that has been a long-standing benchmark for image classification.  
We used a standard ResNet-50 model pretrained on ImageNet and replaced its final fully connected layer with a linear layer of size $(2048, 5)$ to match the number of DR classes.

\paragraph{CoaT-Lite-Medium:} For the modern hybrid architecture, we selected the Co-Scale Conv-Attentional Image Transformer (CoaT) \cite{xu2021co}. CoaT uniquely combines the local feature extraction strengths of convolutions with the global context modeling capabilities of self-attention. By processing image features across multiple interacting scales, it offers a powerful alternative to purely convolutional or transformer-based models. We specifically used the `CoaT-Lite-Medium' variant for its strong balance of performance and computational efficiency.

\subsection{Data Augmentation Strategies}

To investigate the impact on CP, we trained separate instances of each model architecture using five distinct augmentation schemes applied only to the proper training set.

\begin{itemize}
    \item \textbf{None:} Only resizing and normalization were applied. This serves as the baseline.
    \item \textbf{Standard:} A common set of transformations including random horizontal flipping and random adjustments to brightness and contrast.
    \item \textbf{CLAHE:} Contrast Limited Adaptive Histogram Equalization was applied to the luminance channel of images to enhance the visibility of retinal vessels and lesions.
    \item \textbf{Mixup:} This technique creates new training examples by taking a convex combination of pairs of images and their labels, which helps regularize the model and smooth decision boundaries \cite{zhang2018mixupempiricalriskminimization}.
    \item \textbf{CutMix:} A patch from one image is cut and pasted onto another, and the labels are mixed proportionally to the area of the patch. This encourages the model to learn from diverse local regions \cite{yun2019cutmix}.
\end{itemize}

\subsection{Conformal Prediction Framework}

We implemented a split conformal prediction procedure based on the k-Nearest Neighbors (k-NN) nonconformity measure \cite{huang2025confineconformalpredictioninterpretable}.
The framework\footnote{The detailed codebase will be released with the final version} can be grouped into following steps:

\begin{enumerate}
    \item \textbf{Feature Embedding:} For a given trained model (e.g., ResNet-50 with Mixup), we first employ it as a feature extractor. 
    All images from the training, calibration, and test sets are passed through the model, and we extract the output vectors from the penultimate layer—specifically, the global average pooling layer for ResNet-50 and the `norm4' layer for CoaT. 
    This process transforms each image into a high-dimensional feature embedding.
    \item \textbf{Nonconformity Score Calculation:} The nonconformity score for a given image measures how dissimilar it is to its own class compared to other classes. For an image embedding $u$ with a true label $y$, its score $\alpha$ is calculated as:
    \begin{equation}
        \alpha(u, y) = \frac{\text{avg\_dist}(u, k\text{-NN of class } y)}{\text{avg\_dist}(u, k\text{-NN of classes } \neq y)}
    \end{equation}
    where `\text{avg\_dist}' is the average cosine distance to the k-nearest neighbors found within the proper training set embeddings. A low score indicates conformity (the image is typical), while a high score indicates nonconformity (the image is atypical).
    \item \textbf{Calibration:} We calculate the nonconformity score for every image in the calibration set using its true label. This provides a reference distribution of scores, establishing what is "normal" for the given model and data.
    \item \textbf{P-Value Generation:} For each image in the test set, we calculate a p-value for each of the five possible DR grades. To get the p-value for a potential grade $\hat{y}$, we temporarily assume it is the true label, calculate a nonconformity score, and then compare this score to the distribution of calibration scores. The p-value is the fraction of calibration scores that are greater than or equal to the test image's score.
    \item \textbf{Prediction Set Formation:} Finally, for a user-chosen error rate $\epsilon$ (e.g., 0.1 for a 90\% guarantee), the prediction set $\Gamma^{\epsilon}$ for a test image is formed by including all DR grades whose p-value is greater than $\epsilon$.
\end{enumerate}

\subsection{Evaluation Metrics}
To assess the impact of augmentation on CP, we used the following metrics:
\begin{itemize}
    \item \textbf{Top-1 Accuracy:} The standard classification accuracy of the model, considering only the single label with the highest p-value. This metric serves as a baseline for the model's raw predictive performance, similar to conventional classification.

    \item \textbf{Marginal Coverage:} The proportion of test images for which the true label is included in the prediction set $\Gamma^{\epsilon}$. A valid conformal predictor must have a coverage $\ge 1-\epsilon$.

    \item \textbf{Average Set Size:} The average number of labels included in the prediction sets across all test images. This metric measures the efficiency or sharpness of the predictions. A smaller average set size is highly desirable as it indicates greater model confidence.

    \item \textbf{Correct Efficiency:} The proportion of prediction sets that contain \textit{exactly one} label, which is also the correct one. It quantifies how often the model is both correct and maximally confident.
\end{itemize}

\section{Results and Discussion}
In this section, we present the results examining the role of 
data augmentation in shaping CP performance for diabetic retinopathy grading.
All results are reported with a target coverage of $1-\epsilon = 0.9$.

\begin{figure}[t]
\centering
\includegraphics[width=0.8\textwidth]{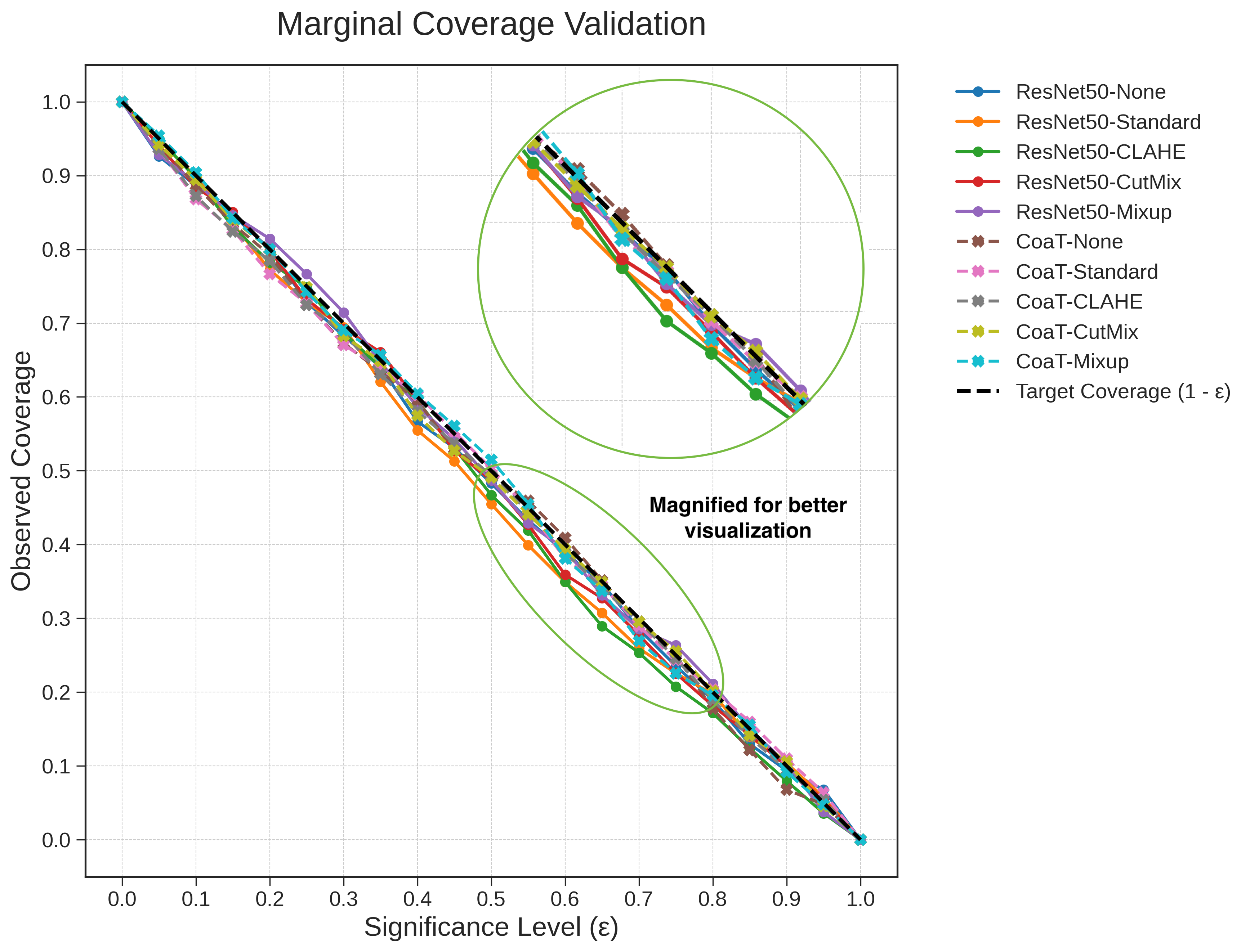}
\caption{Marginal coverage curves for all 10 models. The dashed line represents the ideal $1-\epsilon$ coverage. Models that fall below this line fail to meet the statistical guarantee of CP.} \label{fig:coverage}
\end{figure}

\begin{table}[ht]
\caption{Primary results for both architectures across all augmentation strategies. Metrics are reported at a target coverage of 90\%.}\label{tab:main_results}
\centering
\begin{tabular}{l|cc|cc}
\toprule
\textbf{} & \multicolumn{2}{c|}{\textbf{ResNet-50}} & \multicolumn{2}{c}{\textbf{CoaT-Lite-Medium}} \\
\cmidrule(r){2-3} \cmidrule(l){4-5}
\textbf{Augmentation} & \textbf{Coverage} & \textbf{Top-1 Acc.} & \textbf{Coverage} & \textbf{Top-1 Acc.} \\
\midrule
None & 0.886 & 0.836 & 0.886 & 0.832 \\
Standard & 0.898 & 0.820 & 0.868 & 0.808 \\
CLAHE & 0.896 & 0.778 & 0.872 & 0.806 \\
CutMix & 0.884 & 0.832 & 0.894 & 0.838 \\
Mixup & 0.890 & 0.838 & 0.904 & 0.842 \\
\bottomrule
\end{tabular}
\end{table}

\begin{figure}[ht]
\centering
\includegraphics[width=\textwidth]{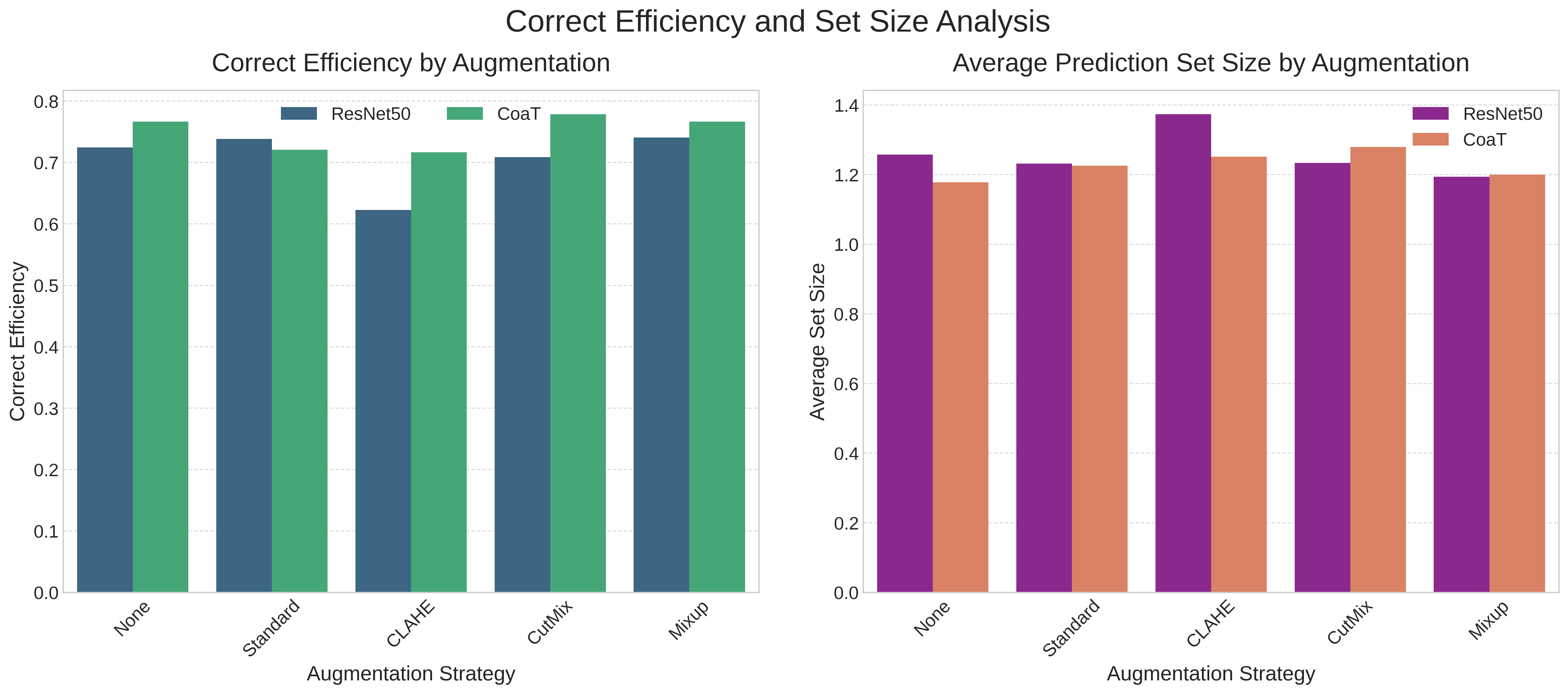}
\caption{Correct efficiency for each model and augmentation strategy. Higher bars indicate a better ability to produce correct, single-label prediction sets.} \label{fig:efficiency_and_set_size}
\end{figure}

We first analyze marginal coverage (Fig. \ref{fig:coverage}), which validates the reliability of the uncertainty estimates. We observe that generally most models track closely with the ideal $1-\epsilon$ coverage line. 
We then present the primary results for both architectures across all augmentation strategies in Table \ref{tab:main_results}, where we analyze coverage and top-1 accuracy. While it is well known that recent augmentation strategies such as CutMix and Mixup improve generalization, we also observe that these advanced methods result in higher coverage. The behavior of other augmentation strategies suggests that they may alter the data distribution in ways that modestly reduce the model's calibration fidelity. Conversely, the CoaT model trained with Mixup is the only configuration to clearly meet the 90\% target level, indicating that this combination of advanced architecture and regularization may best preserve the exchangeability assumption vital for (CP).

In Fig. \ref{fig:efficiency_and_set_size}, we present an analysis of correct efficiency and average prediction set size for each model across different augmentation strategies. First, in terms of correct efficiency (left), the CoaT-Lite-Medium architecture generally outperforms ResNet-50. Interestingly, we also observe that CLAHE—a commonly used enhancement for fundus images—yields the lowest efficiency for ResNet-50 and among the lowest for CoaT. This suggests that while CLAHE may improve visual contrast for human interpretation, it could disrupt feature consistency in a way that compromises model certainty.
This observation is consistent with the average prediction set size (right): for ResNet-50, CLAHE results in the largest average prediction set, implying greater model uncertainty. For CoaT-Lite-Medium, the set sizes are more consistent; however, CutMix produces a slightly larger set, possibly because it encourages the model to learn from more varied and occasionally ambiguous patch combinations. Mixup yields the smallest set size for ResNet-50 and remains competitive for CoaT, highlighting its strength in constraining uncertainty.

\begin{figure}[ht]
\centering
\includegraphics[width=\textwidth]{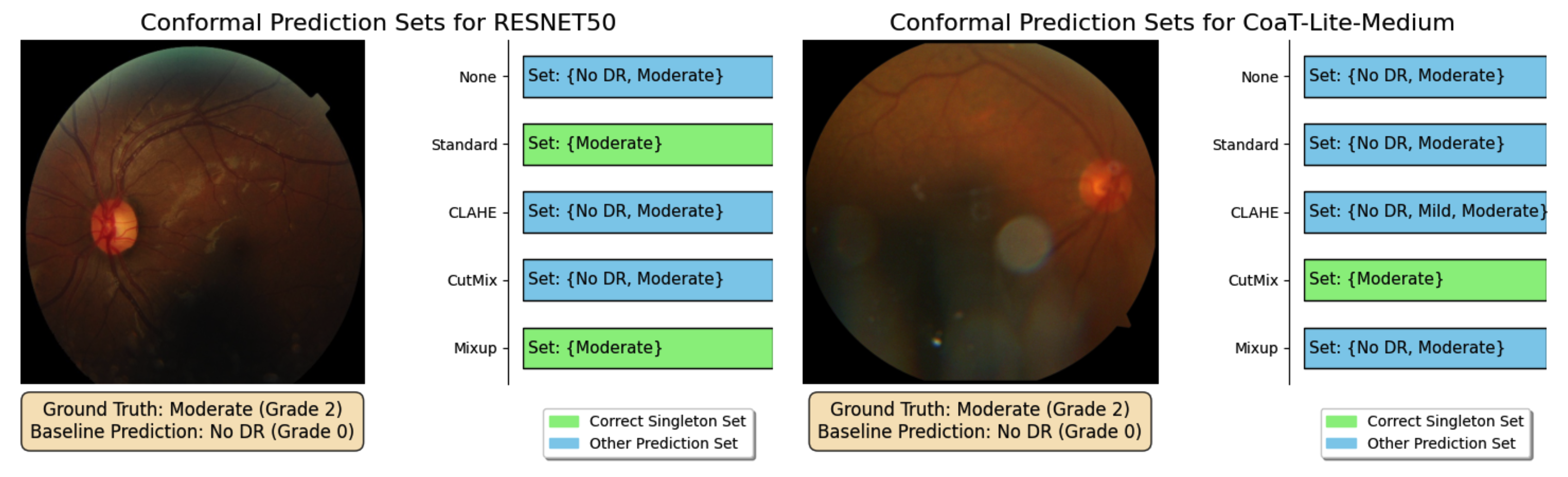}
\caption{Visual illustration of CP in practice. For each model category, two images were randomly selected where the baseline prediction was incorrect. To the right of each image, we show the prediction sets generated under different augmentation strategies.} \label{fig:visual_example}
\end{figure}

Finally, in Fig. \ref{fig:visual_example}, we present a visual illustration of CP on two randomly selected examples where the baseline prediction was incorrect. As shown in both cases, while all prediction sets successfully included the ground truth label (i.e., moderate DR), only a few augmentation strategies produced a correct singleton set.

\section{Conclusion}
In this paper, we conducted a systematic investigation into the interplay between data augmentation and CP for the critical task of diabetic retinopathy grading. Our experiments, which spanned two distinct architectures (ResNet-50 and CoaT-Lite-Medium) and five augmentation strategies, demonstrate that the choice of augmentation profoundly influences not only predictive accuracy but also the validity and efficiency of uncertainty quantification. Advanced strategies such as Mixup and CutMix tend to produce more reliable models—both in terms of raw performance and conformity with prediction guarantees. In contrast, conventional preprocessing methods like CLAHE, while potentially enhancing interpretability for human viewers, may inadvertently undermine model certainty and calibration. These findings underscore a key insight: for the safe and effective deployment of AI in clinical settings, data augmentation should not be optimized solely for accuracy. Instead, it must be co-designed and rigorously evaluated with downstream reliability and trustworthiness in mind.

This work lays a foundation for further research into designing augmentation strategies that not only enhance performance but also uphold statistical guarantees under conformal prediction. Future studies could build on our findings by exploring a wider spectrum of model architectures and validating across diverse datasets to assess generalizability. In addition, deeper analysis into how specific augmentations can shape model certainty could offer valuable intuition for designing more trustworthy pipelines. Ultimately, advancing these directions will be essential for aligning predictive performance with the practical demands of clinical deployment, where reliability, interpretability, and efficiency must go hand in hand.

\section{Acknowledgment}
This work was supported by National Institutes of Health grant P20 GM144230 from the National Institute of General Medical Sciences (NIGMS).

\bibliographystyle{splncs04}
\bibliography{mybibliography}

\begin{thebibliography}{10}
\providecommand{\url}[1]{\texttt{#1}}
\providecommand{\urlprefix}{URL }
\providecommand{\doi}[1]{https://doi.org/#1}

\bibitem{angelopoulos2021gentle}
Angelopoulos, A.N., Bates, S.: A gentle introduction to conformal prediction and distribution-free uncertainty quantification. arXiv preprint arXiv:2107.07511  (2021)

\bibitem{angelopoulos2024conformal}
Angelopoulos, A.N., Pomerantz, S., Do, S., Bates, S., Bridge, C.P., Elton, D.C., Lev, M.H., Gonz{\'a}lez, R.G., Jordan, M.I., Malik, J.: Conformal triage for medical imaging ai deployment. medRxiv pp. 2024--02 (2024)

\bibitem{barber2023conformal}
Barber, R.F., Candes, E.J., Ramdas, A., Tibshirani, R.J.: Conformal prediction beyond exchangeability. The Annals of Statistics  \textbf{51}(2),  816--845 (2023)

\bibitem{gal2016dropout}
Gal, Y., Ghahramani, Z.: Dropout as a bayesian approximation: Representing model uncertainty in deep learning. In: international conference on machine learning. pp. 1050--1059. PMLR (2016)

\bibitem{gulshan2016development}
Gulshan, V., Peng, L., Coram, M., Stumpe, M.C., Wu, D., Narayanaswamy, A., Venugopalan, S., Widner, K., Madams, T., Cuadros, J., et~al.: Development and validation of a deep learning algorithm for detection of diabetic retinopathy in retinal fundus photographs. jama  \textbf{316}(22),  2402--2410 (2016)

\bibitem{he2016deep}
He, K., Zhang, X., Ren, S., Sun, J.: Deep residual learning for image recognition. In: Proceedings of the IEEE conference on computer vision and pattern recognition. pp. 770--778 (2016)

\bibitem{huang2025confineconformalpredictioninterpretable}
Huang, L., Lala, S., Jha, N.K.: Confine: Conformal prediction for interpretable neural networks (2025), \url{https://arxiv.org/abs/2406.00539}

\bibitem{huang2023identifying}
Huang, Y., Lin, L., Cheng, P., Lyu, J., Tam, R., Tang, X.: Identifying the key components in resnet-50 for diabetic retinopathy grading from fundus images: a systematic investigation. Diagnostics  \textbf{13}(10), ~1664 (2023)

\bibitem{kelly2019key}
Kelly, C.J., Karthikesalingam, A., Suleyman, M., Corrado, G., King, D.: Key challenges for delivering clinical impact with artificial intelligence. BMC medicine  \textbf{17}(1), ~195 (2019)

\bibitem{lakshminarayanan2017simple}
Lakshminarayanan, B., Pritzel, A., Blundell, C.: Simple and scalable predictive uncertainty estimation using deep ensembles. Advances in neural information processing systems  \textbf{30} (2017)

\bibitem{li2019diagnostic}
Li, T., Gao, Y., Wang, K., Guo, S., Liu, H., Kang, H.: Diagnostic assessment of deep learning algorithms for diabetic retinopathy screening. Information Sciences  \textbf{501},  511--522 (2019)

\bibitem{liu2024lesion}
Liu, C., Wang, W., Lian, J., Jiao, W.: Lesion classification and diabetic retinopathy grading by integrating softmax and pooling operators into vision transformer. Frontiers in Public Health  \textbf{Volume 12 - 2024} (2025). \doi{10.3389/fpubh.2024.1442114}

\bibitem{shafer2008tutorial}
Shafer, G., Vovk, V.: A tutorial on conformal prediction. Journal of Machine Learning Research  \textbf{9}(3) (2008)

\bibitem{shorten2019survey}
Shorten, C., Khoshgoftaar, T.M.: A survey on image data augmentation for deep learning. Journal of big data  \textbf{6}(1),  1--48 (2019)

\bibitem{smucny2022data}
Smucny, J., Shi, G., Lesh, T.A., Carter, C.S., Davidson, I.: Data augmentation with mixup: Enhancing performance of a functional neuroimaging-based prognostic deep learning classifier in recent onset psychosis. NeuroImage: Clinical  \textbf{36},  103214 (2022)

\bibitem{srilakshmi2025data}
Srilakshmi, U., Kumar, K.V., Korimilli, S., Goutham, S., Golamari, J.M., Brundavani, P.: Data augmentation-based diabetic retinopathy classification and grading with the dynamic weighted optimization approach. Special Issue: Technologies and Their Effects on Real-Time Social Development  \textbf{167}(A2 (S)) (2025)

\bibitem{ting2016diabetic}
Ting, D.S.W., Cheung, G.C.M., Wong, T.Y.: Diabetic retinopathy: global prevalence, major risk factors, screening practices and public health challenges: a review. Clinical \& experimental ophthalmology  \textbf{44}(4),  260--277 (2016)

\bibitem{vazquez2022conformal}
Vazquez, J., Facelli, J.C.: Conformal prediction in clinical medical sciences. Journal of Healthcare Informatics Research  \textbf{6}(3),  241--252 (2022)

\bibitem{vovk2005algorithmic}
Vovk, V., Gammerman, A., Shafer, G.: Algorithmic learning in a random world. Springer (2005)

\bibitem{xu2021co}
Xu, W., Xu, Y., Chang, T., Tu, Z.: Co-scale conv-attentional image transformers. In: Proceedings of the IEEE/CVF international conference on computer vision. pp. 9981--9990 (2021)

\bibitem{yun2019cutmix}
Yun, S., Han, D., Oh, S.J., Chun, S., Choe, J., Yoo, Y.: Cutmix: Regularization strategy to train strong classifiers with localizable features. In: Proceedings of the IEEE/CVF international conference on computer vision. pp. 6023--6032 (2019)

\bibitem{zhang2018mixupempiricalriskminimization}
Zhang, H., Cisse, M., Dauphin, Y.N., Lopez-Paz, D.: mixup: Beyond empirical risk minimization (2018), \url{https://arxiv.org/abs/1710.09412}

\bibitem{zhou2025conformal}
Zhou, X., Chen, B., Gui, Y., Cheng, L.: Conformal prediction: A data perspective. ACM Comput. Surv.  (May 2025). \doi{10.1145/3736575}, \url{https://doi.org/10.1145/3736575}, just Accepted

\end{thebibliography}
\end{document}